\documentclass[final,3p,times,twocolumn]{elsarticle}

\usepackage{epsfig}
\usepackage{amsmath}
\usepackage{multirow}
\usepackage[ruled,vlined]{algorithm2e}
\usepackage{amssymb}
\usepackage{amsthm}
\usepackage{balance}

\RequirePackage{filecontents}

\journal{Journal of Information Sciences}

\begin{document}

\begin{frontmatter}



\title{Cross-Subject Transfer Learning in Human Activity Recognition Systems using Generative Adversarial Networks}


\author[mymainaddress]{Elnaz Soleimani}
\ead{elnaz.soleimani@aut.ac.ir}
\author[mysecondaryaddress]{Ehsan Nazerfard\corref{mycorrespondingauthor}}
\cortext[mycorrespondingauthor]{Corresponding author}
\ead{nazerfard@aut.ac.ir}

\address[mymainaddress]{Ambient
Intelligence Research Lab, Department of Computer Engineering and Information Technology, Amirkabir University of Technology, 424 Hafez Ave. Tehran, Iran.}
\address[mysecondaryaddress]{Department of Computer Engineering and Information Technology, Amirkabir University of Technology, 424 Hafez Ave. Tehran, Iran.}

\begin{abstract}
 Application of intelligent systems especially in smart homes and health-related topics has been drawing more attention in the last decades. Training Human Activity Recognition (HAR) models - as a major module- requires a fair amount of labeled data. Despite training with large datasets, most of the existing models will face a dramatic performance drop when they are tested against unseen data from new users. Moreover, recording enough data for each new user is unviable due to the limitations and challenges of working with human users. Transfer learning techniques aim to transfer the knowledge which has been learned from the source domain (subject) to the target domain in order to decrease the models' performance loss in the target domain. This paper presents a novel method of adversarial knowledge transfer named SA-GAN stands for \textbf{S}ubject \textbf{A}daptor GAN which utilizes \textbf{G}enerative \textbf{A}dversarial \textbf{N}etwork framework to perform cross-subject transfer learning in the domain of wearable sensor-based Human Activity Recognition. SA-GAN outperformed other state-of-the-art methods in more than 66\% of experiments and showed the second best performance in the remaining 25\% of experiments. In some cases, it reached up to 90\% of the accuracy which can be obtained by supervised training over the same domain data. \end{abstract}


\end{frontmatter}


\section{Introduction}
\label{intro}
Remarkable enrichment of sensor technology and consequently smart environments alongside with huge progress in machine learning techniques have pervasively brought intelligent solutions into every aspect of human life. Recognition of what the human subject is doing, widely considered to be one of the most important tasks of an intelligent system known as an active field of research named Human Activity Recognition (HAR).

Previous studies on HAR can be generally categorized based on sensor modalities and data utilized for detection of activity details include vision and sensors based approaches. Vision-based sensors are exploited to capture images, videos or surveillance camera features to recognize activity\cite{nweke2018deep}. Despite the successful performance of vision based solutions, non-visual sensors are still required to address their existing limitations such as laborious processing and privacy problems. Non-visual sensors can be installed on the human's body (wearable sensors) or in the environment (ambient sensors). Utilizing a network of heterogeneous sensors has become widespread interest as well.

Diverse supervised and semi-supervised machine learning models have been proposed for activity recognition. These models deliver promising accuracy conditioning on training with enough labeled data. However, the pitfall is that their performance will dramatically fall against data, from new unseen distributions. The difference may root in feature space or label space distribution. Therefore, recognizing activities of a new user remains challenging for the model which was trained by samples of other users' behavior. Nevertheless, collecting and labeling sufficient training data is not feasible for every new user since it requires a relatively long time observation of human subjects behavior which is time-consuming and sometimes impractical.

Transfer learning techniques aim to prevent that performance leak by adapting obtained knowledge from the source domain (training users) to the target domain (new users). Transfer learning is researched under a variety of different names such as life-long learning, knowledge transfer, learning to learn, inductive transfer, context-sensitive learning, and meta-learning in the field of machine learning,\cite{cook2013transfer}.

This research study investigates how to solve the aforementioned limitations and analyzes the results of our proposed solution. The remainder of this paper is organized as follows in five sections:

Section \ref{related_work} examines the previous research works have been devoted to the study of HAR and Transfer Learning. Section \ref{proposed_model} describes SA-GAN and its related training details. Evaluation, experimental results and their analysis are presented in section \ref{experiments}. Finally, section \ref{future_works} summarizes the results of this work, draws conclusions and highlights issues for future researches.

\section{Related Works}
\label{related_work}
\subsection{Human Activity Recognition}
Traditional machine learning approaches including K-Nearest Neighbor(KNN), Hidden Markov Model (HMM), Support Vector Machine (SVM), Random Forest (RF) and Naive Bayes have shown satisfactory results on recognizing human activities \cite{mannini2010machine,bedogni2012train}. A major criticism of these models is that they mainly rely on handcrafted feature extraction or heuristic information. Besides the demand for the domain specialist, extracted features are not abstracted enough. Therefore, models are not suitable for generalization and recognition of more abstract activities\cite{yang2009activity}.

Deep learning approaches have been interestingly used in feature extraction applications such as HAR problems which deal with high dimensional data \cite{vepakomma2015wristocracy,ronao2016human}. Data-driven \cite{cheng2017human,chen2015deep, yang2015deep} and Model-driven approaches \cite{ha2015multi, jiang2015human} are two primary ways of deep model application in HAR problems. Increasing network's depth improves the quality of extracted features\cite{wang2018deep}.

Stacked Auto Encoders (SAE) are the models capable of learning lower dimensional representation in an unsupervised manner \cite{li2014unsupervised}. Recurrent networks and their combination with Restricted Boltzmann Machines would be of interest owing to the temporal nature of human activities data \cite{jiang2015human,guan2017ensembles,inoue2018deep}. Nonetheless, their high resource consumption and low rate of learning is counted as their prohibitive drawbacks.

\subsection{Knowledge Transfer}
The literature on knowledge transfer can be generally categorized into four main approaches based on the type of knowledge they transfer \cite{pan2010survey}:
\begin{itemize}
  \item Instance Transfer:
    Methods placing in this approach, mainly aim for weighting and transforming labeled instances into the target domain. Standard supervised machine learning models can be applied on transferred samples afterward.

  \item Feature Representation Transfer:
  The core idea of this category's models is about finding a common representation of both source and target domain that decreases the distance between domains while keeping their classes discernible.

  \item Parameter Transfer:
  The basic assumption is that the source and target domains share some parameters or prior distributions of the models' hyper-parameters. These methods focus on the transformation of prior knowledge and parameters between domains.

  \item Relational Transfer:
    The knowledge to be transferred is the relationship among the data. Mapping of relational knowledge between the source domain and target domains is built. Both domains should be relational.

\end{itemize}

Authors in \cite{hu2011cross} proposed an instance-based transfer model in HAR domain that interprets the data of source domain as pseudo training data with respect to their similarity measure to the target domain samples. These pseudo data then will be fed into supervised learning algorithms for training the classifier.

Quite recently, another Cross-Domain Activity Recognition translation framework was proposed by researchers in \cite{wang2018stratified}. It first obtains pseudo labels for the target domain using the majority voting technique. Then, it transforms both domains into common subspaces considering intra-class correlations. This model which is working in a semi-supervised manner obtains labels of target domain via the second annotation.

Transfer Component Analysis (TCA) is a domain adaptation method introduced in \cite{pan2011domain}. TCA learns transfer components across domains in a Reproducing Kernel Hilbert Space for establishing a representation transfer. With the new representation in the subspace spanned by these transfer components, standard machine learning methods are applicable to train classifiers or regression models in the source domain for use in the target domain.

Another representation transfer solution is described in \cite{gong2012geodesic} named in short GFK which is a kernel-based method. It models the domain shift by integrating an infinite number of subspaces that characterize changes in statistical and geometric attributes from the source to the target domain.

\subsection{Generative Adversarial Networks}
The idea of Adversarial Learning has attracted much attention from research teams since the introduction of the GAN framework. Several publications have appeared in recent years documenting Domain Adaptation using GAN models\cite{luo2017label}. Though, most of the previous works concentrating on utilizing GAN in the domain of Vision and Image Processing.

Authors in \cite{zhu2017unpaired} present an approach for learning to translate an image from a source domain to a target domain in order to overcome the problem of lacking labeled images.
Researchers in \cite{yi2017dualgan}, developed an innovative mechanism named dual-GAN, which provides an image translator trained by sets of unlabeled images of both domains.
A new unsupervised method presented in \cite{bousmalis2017unsupervised} that learns a transformation in the pixel space from one domain to the other by adapting source-domain images to appear as if drawn from the target domain. Isola et al. \cite{isola2017image} examine conditional adversarial networks as a general-purpose solution for image-to-image translation problems.

However, research on employing GAN models in Human Activity Recognition domain is limited to approaches of generating high-quality artificial data, imitating output of wearable sensors\cite{saeedi2018personalized}. Despite compelling results of GANs on the vision-based problems, they are not optimal on discriminative tasks and can handle smaller domain shifts\cite{tzeng2017adversarial}. To the authors' best knowledge, very few publications can be found in the literature that discusses the issue of Knowledge Transfer using GAN for classification performance improvement in the HAR problems.

\section{Proposed Model: SA-GAN}
\label{proposed_model}
Following our semi-supervised knowledge transfer setting, we have labeled data of the source domain $(X_{s},Y_{s})\sim P_{s}$ and unlabeled data of target domain $ X_{t}\sim P_{t}$. In our Cross-Subject Transfer Learning problem the difference between the domains roots in the distribution of feature space and the conditional distribution of label space. It can be interpreted as the scenario when there is a model trained with limited samples from distribution $P_{s}$, and it is required to test the model against samples drawn from $P_{t}$. Given these assumptions, the goal is to perform transformation of samples from $P_{s}$ in order to have labeled data as if drawn from $P_{t}$. Concurrently a classifier can be trained with those transferred instances. Therefore it will be able to classify data from $P_{t}$.
\begin{figure*}[t]
    \setlength{\fboxrule}{0pt}
    \framebox{
      \parbox{0.95\textwidth}{
        \centering
        \includegraphics[width=0.95\textwidth, height=10cm, keepaspectratio]{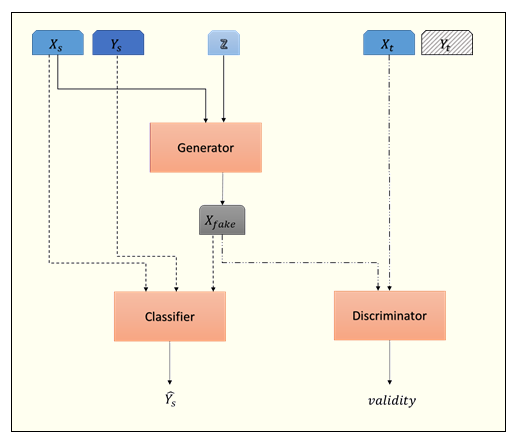}}}
        \caption{Abstract structure of SA-GAN. Dotted, dashed and solid lines depict input data flow to the Discriminator, Classifier, and Generator respectively.}
        \label{ost_structure}
\end{figure*}
As shown in Fig. \ref{ost_structure}, our proposed model consists of three main components:
\begin{itemize}
  \item Generator (G): This component is in charge of generating artificial data which are similar to the data from the target domain.
  \item Discriminator (D): This component's task is to distinguish between artificial G's output and real data from the target domain.
  \item Classifier (C): This component aims to assign a true label to its inputs. Its interaction in training of adversarial components prevents the model from Mode Collapse related challenges.
\end{itemize}

Similar to the classic GAN model, G and D play a minimax game with the value function $\mathcal{V}_{adv}$ as the following:

\begin{equation}
    \begin{aligned}
    \label{eq:eq_adv}
    \mathcal{V}_{adv}(D,G)= E_{x_{t}}\Big[\log ⁡D(x_{t})\Big]+ E_{x_{s}} \Big[\log ⁡    \Big(1-D\big(G(x_{s})\big)\Big) \Big]
    \end{aligned}
\end{equation}

In this game, G implicitly defines a new distribution $P_g$ which supposed to be as close as possible to $P_{t}$ in a way that D will not be able to discriminate data from $P_g$ and $P_t$. This minimax game has a global optimum in $P_g = P_t$; hence the optimal discriminator can be written in the form of\cite{goodfellow2014generative}:

\begin{equation}
\label{eq:eq_optimal}
    D^{*}(x) = \frac{P_{t}(x)}{P_{g}(x) + P_{t}(x)}  = \frac{1}{2}
\end{equation}

Which means $D^{*}$ will discriminate half of the samples incorrectly, as G getting more powerful in generating target-wise data.

\begin{figure}[t]
    \setlength{\fboxrule}{0pt}
    \framebox{
      \parbox{0.45\textwidth}{
        \centering
        \includegraphics[width=0.45\textwidth, height=6cm, keepaspectratio]{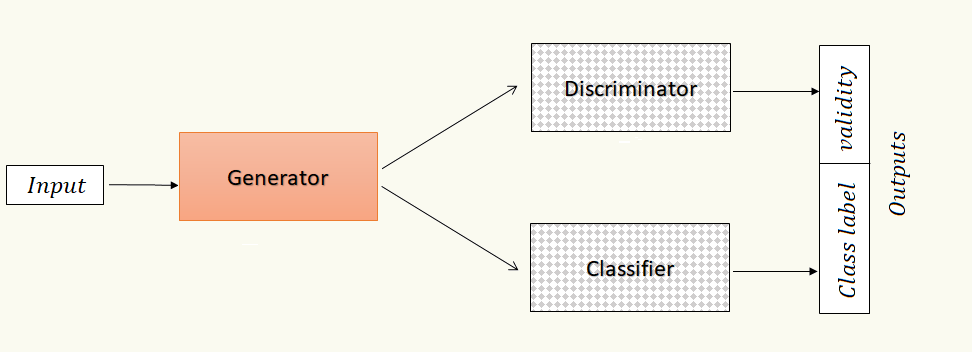}}}
        \caption{Generator gets updated obtaining the gradient from the combination of classifier and discriminator's output.}
        \label{generator_input}
\end{figure}

Simultaneously, classifier C prevents generator G from collapsing since it gets updated based on discernibility of its output evaluated by C, as illustrated in Fig. \ref{generator_input}. Mode collapse refers to an state when G collapses too many values of input($X_{source}$) to the same value of output($X_{fake}$)\cite{goodfellow2014generative}. The equation that describes classifier's supervised optimization function is as follows:
\begin{equation}
\label{eq:eq_cls}
    \mathcal{V}_{cls}(G,C) = E_{(x_{s},y_{s} )} \Big[ -y_{s}  \log⁡\Big(C\big(G(x_{s} )\big)\Big)-y_{s}  \log⁡C(x_{s})\Big]
\end{equation}

Given $\mathcal{V}_{adv}$ and $\mathcal{V}_{cls}$ the overall objective function of the framework is defined by Equation \ref{eq:eq_overal} whereas $\lambda_{adv}$ and $\lambda_{cls}$ are adversarial and classification task factor:
\begin{equation}
\label{eq:eq_overal}
    \min_{G,C} \max_{D}  \lambda_{adv} \mathcal{V}_{adv}(D,G) + \lambda_{cls} \mathcal{V}_{cls}(G,C)
\end{equation}
 These two hyperparameters determine the impact level of D and C's output on gradient update of G. Considering quick convergence of adversarial components, large values of $\lambda_{cls}$ let classifier keep improving using transferred and original target data.

\subsection{Components' Architecture}
Summary of all components' parameters and their input/output are provided in Table \ref{table_parameters}. Those parameters deliver the capability of complexity control to the model. Model's sophistication should suit source and target subject distance so that be able to move the required mass between two distribution. Complexity can be regulated based on the model's fitness reflection in loss values' trend.

For discriminator D, we have implemented a model composed of \textit{$d_{f}$} convolutional layers following by Leaky ReLU. For the first convolutional layer, we also added up Batch Normalization. The last layer of this component is a 1D convolutional layer with the number of features outputs. By practical investigation, we decided to apply \textit{Tanh} as the activation function for the last layer, to obtain $\pm1$ output instead of binary ones. Besides, One-sided label smoothing has been done to make further improvement as is suggested in \cite{salimans2016improved}.

Several experiments had been carried out with the purpose of verifying performance loss without applying transfer learning techniques and the necessity of their application. During these experiments, different architectures were examined. Admissible performance and generalization potential of convolutional models led us to opt for a convolutional architecture for classifier C and generator G. Generator and classifier components, consist of residual blocks with {$g_{f}$} and {$c_{f}$} filters in convolutional layers respectively.

\subsection{Training}
Model training entails three main steps implemented in mini-batch mode.
In each step, only one component is getting updated, and two remaining components remain constant during training batch. Moreover, training steps are considered to be independent optimization problem so that different optimizers, learning rates, and loss functions can be applied to solve the problem. Mini-batch training allowed the model to take into account a bunch of data once in an iteration; so it has a wider horizon which is helpful to generate more various samples. While in the simple stochastic  training the model processes each sample independently and it probably makes the model blind to the diversity of its generated samples. Algorithm \ref{Algorithm} outlines the training procedure.

In the first step, discriminator D is updated by maximizing Equation \ref{eq:eq_adv} using samples from both domains. This step is sort of supervised training that exploits the feature space of $X_{t}$, and $X_{s}$ which has adopted by Generator G, as the input and yields a binary label \textit{validity} which remarks real or fake nature of data. In our implementation, Mean Squared Error(MSE) loss and Stochastic Gradient Descent(SGD) optimizer are chosen for this step.

With the completion of the first step, we can proceed to classifier training based on Equation \ref{eq:eq_cls}. The second step can be treated as a supervised classification problem which aims to assign correct label $Y_{t}$ to both inputs $X_{t}$ and artificial data $X_{fake}$ generated by G using $X_{s}$. Note that the higher objective is to transfer data from source domain through the generator in a way that we get $P_{g}$ close enough to $P_{t}$  so as to be appropriate for classifier training.

Having D and C updated, the final adjustment is to train the generator. As illustrated in Fig. \ref{generator_input}, this element utilizes the combination of discriminator and classifier's output in order to compute its gradient. Each output participates in the training procedure proportional to their task factor as formulated in Equation \ref{eq:eq_overal}.

\begin{algorithm*}[ht]
\caption{
    {Mini-batch stochastic gradient descent training of the proposed model. The mini-batch size, m, $\lambda_{adv}$, and $ \lambda_{cls}$ are hyperparameters. Different optimizers can be used in step (3) to (5).}
}
\KwIn{$X_{s}, X_{t}, Y_{s}$}
\KwOut{C}
\BlankLine
Make random mini-batches of inputs with size m.
\\
\For{number of training iterations or until convergence}{
    \For{number of mini-batches of data}{
    \begin{enumerate}

      \item Sample a mini-batch of $X_{s}, X_{t}, Y_{s}$ from data:
        $\{x_{s}^{(1)}, x_{s}^{(2)}, ... , x_{s}^{(m)}\},
         \{y_{s}^{(1)}, y_{s}^{(2)}, ... , y_{s}^{(m)}\},
         \{x_{t}^{(1)}, x_{t}^{(2)}, ... , x_{t}^{(m)}\}$
      \item  Sample a mini-batch of noise samples $\{z^{(1)},z^{(2)}, ... , z^{(m)}\}$
      \item  Update discriminator D by ascending its stochastic gradient:\\
          \begin{center}
             $\nabla_{\theta_{D}} \mathlarger{ \frac{1}{m} \sum_{i=1}^{m} }\Big[ \log D\big( x_{t}^{(i)}\big) + \log \Big( 1- D\big(G(x_{s}^{(i)}, z^{(i)})\big) \Big) \Big]$
          \end{center}

      \item Update classifier C by ascending its stochastic gradient:\\
          \begin{center}
             $\nabla_{\theta_{C}} \mathlarger{ \frac{1}{m} \sum_{i=1}^{m}} \Big[  -y_{s}^{(i)} \log C\big( x_{s}^{(i)}\big) - y_{s}^{(i)}\log \Big( C\big ( G(x_{s}^{(i)}, z^{(i)} ) \big) \Big) \Big]$
          \end{center}

      \item Update generator G by ascending its stochastic gradient:\\

             $\nabla_{\theta_{G}} \mathlarger{ \frac{1}{m} \sum_{i=1}^{m}}\Bigg[
             \lambda_{cls} \Bigg(  -y_{s}^{(i)} \log C\big( x_{s}^{(i)}\big)- y_{s}^{(i)}\log \Big( C\big ( G(x_{s}^{(i)}, z^{(i)} ) \big) \Big) \Bigg) + \lambda_{adv}  \Bigg(\log D\big( x_{t}^{(i)}\big) + \log \Big( 1- D\big(G(x_{s}^{(i)}, z^{(i)})\big) \Big)  \Bigg)
             \Bigg]$

      \end{enumerate}
    }
}
Return classifier C.

\label{Algorithm}
\end{algorithm*}

\begin{table}
\begin{center}
\caption{Summary of the SA-GAN's elements. \textit{Parameters} column demonstrates complexity controlling constraints for tuning each component to achieve overall equilibrium. $X_{fake}$ is Generator's output by processing $X_{t}$.}
\label{table_parameters}
\centering
\resizebox{\columnwidth}{!}{
\begin{tabular}{@{\extracolsep{\fill}}c|ccc}
Element          & Input    & Output   & Parameters     \\
\hline \hline \\[-1.5ex]
D          & $X \in \{ X_{t} \cup X_{fake}\}$   & $validity \in \{-1,0.9\}$    & $d_{f}$   \\
 & & & \\

C          & $X \in \{ X_{s} \cup X_{fake}\}$   & $Y_{s}$                       & $ c_{f},\lambda_{cls} $   \\
\multirow{2}*{G} &  \multirow{2}*{$X = X_{s} + z $} &  \multirow{2}*{$[validity,Y_{s}]$}  & $g_{f},\lambda_{adv},$ \\
                                                          & & & $\#blocks$
\end{tabular}
}
\end{center}

\end{table}

\section{Experiments}
\label{experiments}
Our experiments are broken down into two groups on the basis of their objectives. The first set of analysis was carried out in order to justify the necessity of applying knowledge transfer technique by investigating the performance drop in case of domain shifts. Another group of experiments was conducted with the aim of measuring the improvement achieved by SA-GAN model.

The evaluation is performed on the Opportunity Challenge benchmark dataset which contains the recorded output of wearable sensors worn by 4 human subjects while they were doing predefined activities\cite{chavarriaga2013opportunity}. There exist three types of activity in this dataset on the basis of their level of abstraction. Recognition task is more difficult for activity with a higher level of abstraction. The most abstract activities, have been picked to evaluate the proposed model.

\begin{figure}[t]
    \setlength{\fboxrule}{0pt}
    \framebox{
      \parbox{0.45\textwidth}{
        \centering
        \includegraphics[width=0.45\textwidth]{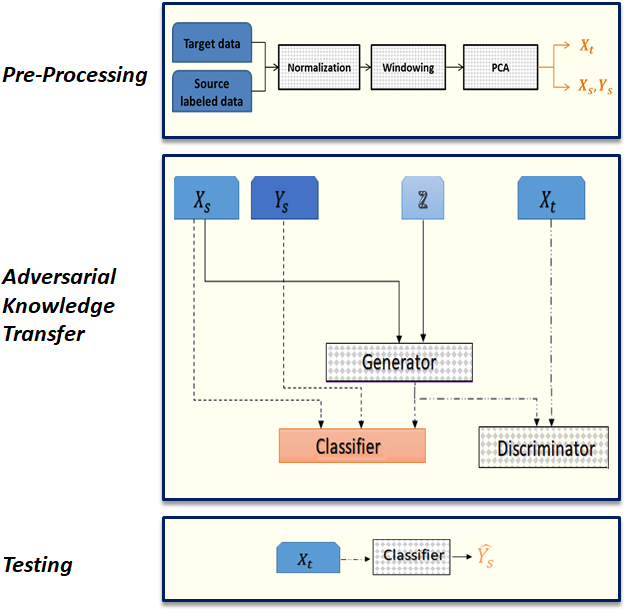}}}
        \caption{Overview of applying the proposed approach. The output of each step is colored in orange.}
        \label{overview}
\end{figure}

For each subject, the first three Activity of Daily Living (ADL) files considered as a training set and fourth and fifth ones were selected as validation and test set respectively. Fig. \ref{overview},  depicts a simple overview of the required steps to have the prediction of target domain $\hat{Y_{t}}$, using SA-GAN.

\subsection{Experimental Setup}

Data preparation is an inevitable step in neural networks training. Our preprocessing framework is composed of three major steps as following:
\begin{enumerate}
  \item Data Preprocessing: In the initial step, missing values of the dataset were replaced by the mean value of their corresponding feature. A min-max normalization has been done based on the sensor's range of output.
  \item Data Segmentation: Approximately 3 seconds length sliding window was applied for segmentation, taking into account 70\% of overlap between successive windows of data.
  \item Dimension Reduction: Considering sliding window application on feature vectors, for each sample we would have around 10 thousands feature values which is extremely difficult for a network to process. Hence, we utilized Principal Component Analysis (PCA) to reduce windows dimension to 88. The number of components to keep, can be determined by a training time versus accuracy trade-off.
\end{enumerate}

\subsection{Results and Analysis}
Our principal objective of transfer learning is to decline the distance between the source and target domain distributions. It is expected to have a more laborious transformation between further domains in terms of time and resource consumption. The distance measurements were taken using Wasserstein distance function which is defined as follows\cite{arjovsky2017wasserstein}:

\begin{equation}
\label{eq:eq_wass}
W(P_{s},P_{t} )=\inf_{\gamma \in \Pi(P_{s}, P_{t})⁡} E_{(x_{s}, x_{t} ) \sim \gamma} ||x_{s}-x_{t}||
\end{equation}

whereas $\Pi(P_{s}, P_{t})$ is the set of all joint distributions $\gamma(x_{s}, x_{t})$ whose marginals are $P_{s}$ and $P_{t}$ respectively. In fact, $\gamma(x_{s}, x_{t})$ denotes the “mass” required to be transported from domain \textit{s} to \textit{t} in order to transform the distributions $P_{s}$ into the distribution $P_{t}$.

Table \ref{table_results} presents the results obtained from our experiments. Each experiment is defined by specific source and target subject whose Wasserstein distance is stated in the third column. This scenario of experiments follows a notion of real-world application where there is a newcomer user to test the pre-trained HAR model while it has not been among training dataset's users. The best source for knowledge transfer can be found either by distance or by validation measure comparison.

\begin{table*}[ht]
\begin{center}
\caption{Comparison of the SA-GAN's performance and GFK \cite{gong2012geodesic}, STL\cite{wang2018stratified} and KNN+PCA model, in terms of W1-F measure. The most dominant performance in each transformation experiment marked in bold.}
\label{table_results}
\centering
\resizebox{\textwidth}{!}{
\begin{tabular}{ccccccccc}
Source Subject     & Target Subject & Wasserstein Distance & No Transfer & KNN+PCA & GFK  & STL  & SA-GAN & Supervised Learning  \\
\hline \hline \\[-1.5ex]
\multirow{3}{*}{1} & 2     & 46.69     & 0.45    & 0.60        & 0.59          & 0.65   & \textbf{0.73} & 0.75     \\
                   & 3     & 45.10     & 0.27    & 0.36        & 0.43          & 0.37   & \textbf{0.45} & 0.71     \\
                   & 4     & 77.15     & 0.40    & 0.48        & \textbf{0.55} & 0.47   & 0.49          & 0.59     \\
\hline \\[-1.5ex]
\multirow{3}{*}{2} & 1     & 40.47     & 0.48    & 0.56        & \textbf{0.62} & 0.52   & 0.56          & 0.65     \\
                   & 3     & 34.38     & 0.44    & 0.49        & 0.51          & 0.46   & \textbf{0.52} & 0.71     \\
                   & 4     & 72.80     & 0.29    & 0.39        & 0.40          &\textbf{0.46} & 0.39    & 0.59     \\
\hline \\[-1.5ex]
\multirow{3}{*}{3} & 1     & 38.38     & 0.23    &\textbf{0.48}& 0.45          & 0.40 & 0.42            & 0.65     \\
                   & 2     & 37.54     & 0.21    & 0.49        & 0.53          & 0.54 & \textbf{0.61}   & 0.75     \\
                   & 4     & 73.69     & 0.31    & 0.35        & 0.44          & 0.37 & \textbf{0.44 }  & 0.59     \\
\hline \\[-1.5ex]
\multirow{3}{*}{4} & 1     & 73.53     & 0.26    & 0.51        & 0.51          & 0.38 & \textbf{0.51}   & 0.65     \\
                   & 2     & 70.80     & 0.29    & 0.48        & 0.45          & 0.54 & \textbf{0.55}   & 0.75     \\
                   & 3     & 69.44     & 0.24    & 0.42        & 0.37          & 0.48 & \textbf{0.49}   & 0.71     \\

\end{tabular}
}
\end{center}

\end{table*}

\begin{table*}[ht]
\begin{center}
\caption{Left: Confusion matrix of the proposed model transferring knowledge from Subject 1 to Subject 2.
Center: Confusion matrix of a supervisely trained model on Subject 2. Right: Confusion matrix of a supervisely trained model on data Subject 1. All the models are tested against Subject 2. }
\label{table_confmat}
\centering
\resizebox{\textwidth}{!}{
\begin{tabular}{l|llllll||llllll||llllll}
 & \multicolumn{6}{|c||}{Transferred from S1 to S2} & \multicolumn{6}{c||}{Supervised Learning on S2} & \multicolumn{6}{c}{Without Transferred}  \\
\hline \hline
    & C0  & C1  & C2  & C3  & C4  & C5     & C0  & C1  & C2  & C3  & C4  & C5   & C0  & C1  & C2  & C3  & C4  & C5\\
\hline
C0  & 257 & 0   & 7   & 19  & 6   & 3      & 198 & 3   & 22  & 24  & 30  & 15   & 201  & 0   & 2   & 1  & 87   & 1  \\
C1  & 15  & 135 & 0   & 12  & 1   & 27     & 1   & 183 & 5   & 1   & 0   & 0    & 8    & 164 & 7   & 2  & 4    & 5  \\
C2  & 13  & 0   & 145 & 16  & 11  & 40     & 2   & 0   & 86  & 34  & 84  & 19   & 20   & 0   & 60  & 11 & 108  & 26 \\
C3  & 62  & 1   & 2   & 256 & 25  & 26     & 6   & 0   & 8   & 301 & 25  & 32   & 113  & 0   & 14  &120 & 104  & 21 \\
C4  & 40  & 0   & 17  & 2   & 109 & 56     & 2   & 0   & 37  & 16  & 123 & 46   & 11   & 2   & 5   & 4  & 142  & 60 \\
C5  & 18  & 8   & 74  & 18  & 40  & 521    & 2   & 0   & 107 & 45  & 49  & 476  & 29   & 4   & 130 & 11 & 200  & 305 \\

\end{tabular}
}
\end{center}
\end{table*}

To assess our proposed model, we have compared its performance with two state-of-the-art transfer learning models including GFK \cite{gong2012geodesic}, STL\cite{wang2018stratified} and a classic knowledge transfer model KNN+PCA in terms of Weighted F1-measure. The most dominant performance of each experiment is in bold font.

On combining models' performance result with domain distances, we deduced that the domains with the largest distance lead to more ineffective transports, as it was expected.
It can be inferred from the reported W-F1 measures that our proposed model made improvement in all the cases compare with No Transfer mode reported in Table \ref{table_results}. Moreover, it shows the predominant results among more than 66\% of experiments and second best classification performance in the remaining 25\% of them.

Further analysis notably showed that in 3 experiments, SA-GAN model has reached up to 90\% accuracy of the model which has been supervisely trained by the target domain labeled data. Supervised learning performance can be assumed as a summit to comprehend how much improvement is feasible to achieve. Therefore, the fourth and last columns of Table \ref{table_results} is a sort of boundary for performance drop and enhancement respectively. Our investigations have shown from 22\% to 47\% of performance loss in No transfer mode.

Table \ref{table_confmat} extends our knowledge of what has been reached by transfer learning. It goes deeper into the confusion matrix of 3 models which are representing Semi-supervised transfer learning (our proposed model), Supervised learning, and No transfer mode. The source domain is Subject 1, and the target is Subject 2. Apart from this slight sign of mode collapse on class 0 (Relaxing) and 5 (Sandwich Time), the result shows appreciable enrichment provided by SA-GAN.

From Table \ref{table_results} it can be seen that performance falls to 0.45 in case of assessing a model that has been trained by samples of Subject 1, against samples of Subject 2 and it heads up to 0.75 given supervised training using Subject 2's labeled data; while the proposed model achieves W-F1 equal to 0.73 which is almost equivalent of supervised learning performance.

\begin{figure}[t]
    \setlength{\fboxrule}{0pt}
    \framebox{
      \parbox{0.45\textwidth}{
        \centering
        \includegraphics[width=0.45\textwidth]{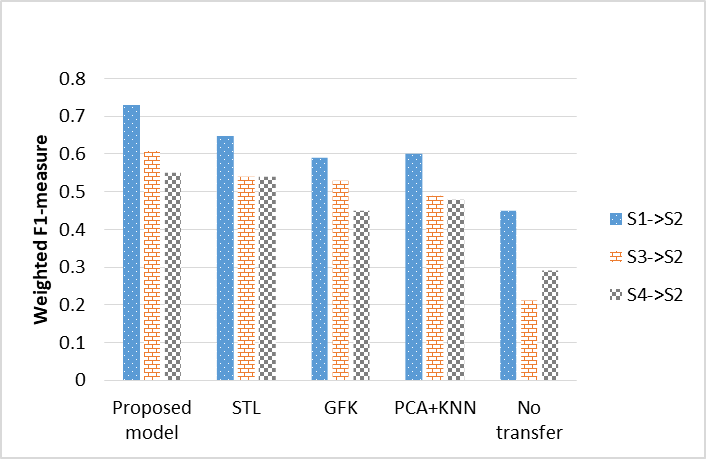}}}
        \caption{Comparison of Knowledge Transfer's results to Subject 2 from different sources by various models.}
        \label{S2_transformatin_result}
\end{figure}

As mentioned earlier, Cross-Subject Transfer Learning using different sources has great potential for practical applications. One typical real-world application is the scenario simulating a situation when a pre-trained machine learning model is facing with a new user, while it is not possible to collect and label enough data for re-training of the model. Assume the model is trained using samples of Subject 1, 3 and 4 and the unseen samples belong to the activities of Subject 2.
For each Knowledge Transfer model, three transformations have been done, using three different sources. These cases are represented in Fig. \ref{S2_transformatin_result}.

Accordingly, Subject 1 is of direct practical relevance and proper to be picked as the source of transfer. Considering Subject 1 to Subject 2 transformation, our proposed model overwhelmed other methods. Summing up the results, for each target subject, most appropriate source of transfer can be found by evaluating the obtained transferred model over a validation set.

\section{Conclusions and Future Work}
\label{future_works}
One of the most important limitations of HAR models lies in lacking a sufficient amount of labeled data. Furthermore, the discovered patterns through available labeled data might not be well generalizable to the samples from unseen subjects. However, data acquisition and labeling are not feasible for newcomers due to limitations of interaction with human users. This paper has highlighted an innovative cutting-edge solution for cross-subject knowledge transfer in the domain of Human Activity Recognition based on Generative Adversarial Network framework. SA-GAN performs a semi-supervised instance-based transfer in order to provide enough data to train a classifier on the target domain. Results so far have been very promising and we reached up to 90\% of supervised model's performance in some cases.

Future work will concentrate on utilizing more stable versions of GAN to prevent mode collapse problem and achieve enhancement on recognition results. To further our research we intend to examine multiple source transfer and combination of transferred models from different source domains.

\bibliographystyle{elsarticle-num}
\balance
\bibliography{mybibfile}

\end{document}